%% file: main.tex
\documentclass[letterpaper]{article} 
\usepackage{aaai2026}  
\usepackage{times}  
\usepackage{helvet}  
\usepackage{courier}  
\usepackage[hyphens]{url}  
\usepackage{graphicx} 
\usepackage{natbib}  
\usepackage{caption} 
\usepackage{algorithm}
\usepackage{algorithmic}
\usepackage{tabularx}
\usepackage{booktabs}
\usepackage{threeparttable}
\usepackage{subcaption}
\usepackage{lscape}
\usepackage{multirow}
\usepackage{newfloat}
\usepackage{listings}
\DeclareCaptionStyle{ruled}{labelfont=normalfont,labelsep=colon,strut=off} 
\floatstyle{ruled}
\newfloat{listing}{tb}{lst}{}
\floatname{listing}{Listing}
\nocopyright

\title{Digital Scale: Open-Source On-Device BMI Estimation from Smartphone Camera Images Trained on a Large-Scale Real-World Dataset}
\author{
    Frederik Rajiv Manichand\textsuperscript{\rm 1,3}, 
    Robin Deuber\textsuperscript{\rm 1}, 
    Robert Jakob\textsuperscript{\rm 1}, 
    Steve Swerling\textsuperscript{\rm 3}, 
    Jamie Rosen\textsuperscript{\rm 3},\\
    Elgar Fleisch\textsuperscript{\rm 1,2}, 
    Patrick Langer\textsuperscript{\rm 1}
}
\affiliations{
    \textsuperscript{\rm 1}Centre for Digital Health Interventions, ETH Zurich\\
    \textsuperscript{\rm 2}Centre for Digital Health Interventions, University of St. Gallen\\
    \textsuperscript{\rm 3}WayBetter\\
    {\tt\small rmanichand@ethz.ch, rdeuber@ethz.ch, rjakob@ethz.ch, steve@waybetter.app, jamie@waybetter.app, efleisch@ethz.ch, planger@ethz.ch}
}

\begin{document}

\maketitle

\input{00_abstract}

\begin{links}
    \link{Code}{https://github.com/im-ethz/DigitalScale}
\end{links}
\input{01_intro}
\input{02_related}

\input{03_method}
\input{04_experiments}

\input{05_discussion}
{\small
  \bibliography{11_references}

}

\input{12_appendix}

\end{document}

%% file: 00_abstract.tex
\begin{abstract}
Estimating Body Mass Index~(BMI) from camera images with machine learning models enables rapid weight assessment when traditional methods are unavailable or impractical, such as in telehealth or emergency scenarios. Existing computer vision approaches have been limited to datasets of up to 14,500 images. In this study, we present a deep learning-based BMI estimation method trained on our WayBED dataset, a large proprietary collection of 84,963 smartphone images from 25,353 individuals. We introduce an automatic filtering method that uses posture clustering and person detection to curate the dataset by removing low-quality images, such as those with atypical postures or incomplete views. This process retained 71,322 high-quality images suitable for training. We achieve a Mean Absolute Percentage Error (MAPE) of 7.9\% on our hold-out test set (WayBED data) using full-body images, the lowest value in the published literature to the best of our knowledge. Further, we achieve a MAPE of 13\% on the completely unseen~(during training) VisualBodyToBMI dataset, comparable with state-of-the-art approaches trained on it, demonstrating robust generalization.  Lastly, we fine-tune our model on VisualBodyToBMI and achieve a MAPE of 8.56\%, the lowest reported value on this dataset so far. We deploy the full pipeline, including image filtering and BMI estimation, on Android devices using the CLAID framework. We release our complete code for model training, filtering, and the CLAID package for mobile deployment as open-source contributions.
\end{abstract}

%% file: 01_intro.tex
\section{Introduction}
\label{sec:intro}

A high Body Mass Index (BMI) is associated with an increased risk of chronic conditions, including cardiovascular disease, certain cancers, and metabolic disorders such as diabetes~\cite{causal_role_high_bmi_chronic_diseases}. Body weight and its fluctuations serve as important indicators of health status and disease risk. For example, in individuals with overweight, a 5\% reduction in body weight can substantially lower the risk of type 2 diabetes and cardiovascular disease~\cite{hornWhatClinicallyRelevant2022}. 
BMI is calculated using weight and height, typically measured with a scale and stadiometer. However, in situations where such equipment is unavailable or impractical, such as remote settings or emergency care, BMI estimation from smartphone camera images offers a digital, real-time alternative. By analyzing images of individuals, deep learning models can estimate BMI without requiring active user input. This may enable integration into clinical and telehealth services, supporting passive BMI tracking, population-level monitoring, early risk screening, and behavioral health interventions. Such methods may also aid weight-based drug dosing when patients are unresponsive, disoriented, or unable to report their weight, addressing a well-documented need for rapid weight estimation in emergency care ~\cite{weight_estimation_comparison_adults,weight_estimation_comparison_children}.


Previous studies have investigated using deep learning models for estimating body weight from images, yielding promising results. \citet{jinAttentionGuidedDeep2022} reported state-of-the-art performance, achieving a Mean Absolute Percentage Error~(MAPE) of 9.37\% for BMI estimation from full-body images. Other studies have focused on weight estimation using facial images alone~\cite{mirabet-herranzNewInsightsWeight2023,vip_attribute}, reporting a Mean Absolute Error~(MAE) of 2.3.

However, most existing work is limited by relatively small datasets—typically comprising no more than 5,900 full-body images~\cite{visual_body_to_bmi}—and often focuses on a single image perspective per dataset, hindering comparisons across different viewpoints. To the best of our knowledge, no prior study has systematically evaluated BMI estimation from full-body, torso, and facial images using data collected from the same individuals. Furthermore, although previous research has demonstrated the feasibility of image-based BMI estimation, none has presented a fully deployable solution, such as models optimized for smartphone execution or code supporting on-device deployment.


We address these limitations with WayBED~(WayBetter BMI Estimation Dataset), our large-scale, real-world dataset comprising over 84,000 full-body images from 25,353 individuals, collected as part of the WayBetter weight-loss program. Our objective is to develop a robust on-device BMI estimation system—one that generalizes well to unseen individuals, can be fine-tuned to new data distributions, and operates efficiently on mobile devices.

To achieve this, we first identify the most suitable image perspective using WayBED by comparing three distinct viewpoints: full-body, torso-up, and face-only. Based on this comparison, we select the best-performing model and evaluate its generalization on the unseen VisualBodyToBMI~\cite{visual_body_to_bmi} dataset. We further fine-tune the model on VisualBodyToBMI to assess its adaptability to new domains, specifically its performance in out-of-distribution scenarios. The final model is deployed in an Android application developed using the CLAID framework, demonstrating its feasibility for real-world use. All code for dataset preparation, model training, mobile deployment, and integration with CLAID is released as open-source contribution (see Table~\ref{tab:introduction:sources_availability} for an overview of available resources).

\begin{table}[h!t]
\caption{Overview of available resources.}
\label{tab:introduction:sources_availability}
\centering
\footnotesize
\renewcommand*{\arraystretch}{1.1}
\begin{tabularx}{\columnwidth}{l|X}
\hline
\textbf{Languages} & Kotlin, Python \\
\textbf{Dataset} & WayBED dataset, available on request \\
\textbf{ML Models} & PyTorch, ExecuTorch models, trained on public dataset, or on WayBED on request \\
\textbf{License} & Apache 2.0 \\
\textbf{Artifacts} & Code for training and deployment, Android App, CLAID package \\
\textbf{Sources} & \url{https://github.com/im-ethz/DigitalScale} \\
\hline
\end{tabularx}
\end{table}
\vspace{0.5em}

The remainder of this work is structured as follows: Section~\ref{sec:related} reviews related work on BMI and weight estimation from images. Section~\ref{sec:method} outlines our methodology, including dataset curation, model development and deployment. Section~\ref{sec:experiments} presents experimental results and Section~\ref{sec:discussion} discusses the findings and outlines limitations.

%% file: 02_related.tex
\section{Related Work}
\label{sec:related}

\begin{table*}[t]
  \centering
  \caption{Related work and commonly used datasets.}
  \label{tab:related_work}
  \renewcommand{\arraystretch}{1.15}
  {\fontsize{8pt}{10pt}\selectfont
  
  \begin{tabularx}{\linewidth}{l X X}
  \toprule
    Paper & Dataset & Estimation Performance\textsuperscript{1} \\
    \midrule

    \citet{visual_body_to_bmi} & 'VisualBodyToBMI': 5900 images, 2950 subjects & MAE: 3.76 (BMI); MAPE: 12.50\% (BMI) \\

    \citet{altinigneHeightWeightEstimation2020} & 4,400 images sourced from social media & MAE: 9.80 (kg) \\

    \citet{PANTANOWITZ2021100727} & 161 individuals, photos as silhouettes & MAE: 1.20 (BMI) \\

    \citet{Majmudar2022} & 134 photos of individuals taken by smartphone & MAE: 2.16 (body fat\% ) \\

    \citet{jinAttentionGuidedDeep2022} & 'VisualBodyToBMI'  + 4190 social-media images & MAE: 3.03 (BMI) MAPE: 9.37\% (BMI) \\

    \citet{kimMultiViewBodyImageBased2023} & 1,000 individuals (front, side, back) & MAE: 3.12 (BMI) \\

    \citet{celeb_fbi} & Front-facing images of 7,211 celebrities & Accuracy 85.60\% on 5-kg weight buckets \\
    \midrule

    \citet{earliest_work_face_2013} &  Morph-II: 14,500 \cite{morph_database} & MAE: 3.14 (BMI) \\

    \citet{kinect_and_mobile} & 50 individuals in three weight classes & Not reported \\

    \citet{vip_attribute} & VIP-attribute dataset: 1026 images of celebrities & MAE: 2.3 (BMI) \\

    \citet{mirabet-herranzNewInsightsWeight2023} & VIP-attribute dataset + 400 images of prisoners & MAE: 7.54 (kg) \\
    \bottomrule
  \end{tabularx}
  }
  \caption*{\footnotesize
  Related work in the upper part is on full-body images, work in the lower part on facial-only images.\\
  \textsuperscript{1}When multiple performances are reported, the best overall value on the dataset is shown.}
\end{table*}

Prior research has investigated a range of approaches for estimating body weight and BMI using machine learning models applied to images captured from various perspectives. Early work primarily focused on facial images~\cite{earliest_work_face_2013,kinect_and_mobile}, while more recent studies have extended to full-body images~\cite{altinigneHeightWeightEstimation2020,visual_body_to_bmi,jinAttentionGuidedDeep2022,kimMultiViewBodyImageBased2023,celeb_fbi} and images captured via smartphone cameras~\cite{Majmudar2022,Farina2022-eo}.
The literature has employed various target variables, including BMI~\cite{PANTANOWITZ2021100727,jinAttentionGuidedDeep2022,kimMultiViewBodyImageBased2023,celeb_fbi}, body weight~\cite{Majmudar2022,earliest_work_face_2013,mirabet-herranzNewInsightsWeight2023}, and body fat metrics~\cite{Farina2022-eo,Majmudar2022}. Table~\ref{tab:related_work} summarizes selected studies on image-based body weight estimation, including the datasets used and reported estimation performance.

\paragraph{Datasets}
Prior research has employed annotated images from a variety of datasets with differing sizes and characteristics. For full-body images, dataset sizes range from around 150 subjects \cite{Majmudar2022,Farina2022-eo, PANTANOWITZ2021100727} to nearly 6000 subjects \cite{visual_body_to_bmi}. For facial images, dataset sizes range from 50 individuals \cite{kinect_and_mobile} to 14,500 images \cite{earliest_work_face_2013}. Various authors have proposed public benchmark datasets for body weight and BMI estimation, such as the Visual-Body-To-BMI dataset \cite{visual_body_to_bmi} and the Celeb-FBI dataset \cite{celeb_fbi}, both containing full-body images sourced from social media and celebrities, respectively. The VIP-attribute dataset \cite{vip_attribute} proposes a benchmark dataset for facial images, containing images of 1026 celebrities.

\paragraph{Existing machine learning approaches}
Early approaches for BMI estimation from images often relied on manually engineered features to extract relevant information prior to model training. Examples include the use of body keypoints~\cite{visual_body_to_bmi,altinigneHeightWeightEstimation2020}, facial landmarks~\cite{kinect_and_mobile}, and silhouette masks capturing body contours~\cite{PANTANOWITZ2021100727}. These features served as inputs to conventional machine learning models such as Support Vector Machines~\cite{earliest_work_face_2013,visual_body_to_bmi} and random forests~\cite{kinect_and_mobile}.
More recent work has adopted deep learning models to learn features directly from images \cite{jinAttentionGuidedDeep2022,vip_attribute, mirabet-herranzNewInsightsWeight2023}. A variety of Convolutional Neural Network (CNN) architectures have been explored, including ResNet \cite{mirabet-herranzNewInsightsWeight2023,vip_attribute} and VGG \cite{celeb_fbi, PANTANOWITZ2021100727}. \citet{jinAttentionGuidedDeep2022} evaluated multiple CNN architectures on the Visual-Body-To-BMI dataset~\cite{visual_body_to_bmi}, identifying a DenseNet architecture that outperformed ResNet and VGG models of comparable size. Their work also introduced an attention mechanism to focus on the most informative regions of the image, further enhancing model performance.

\paragraph{Full-body and face-only perspectives}
Prior work has investigated various perspectives of images for BMI estimation. The primary distinction has been made between full-body images and facial images. 
Using full-body images from the VisualBodyToBMI dataset~\cite{visual_body_to_bmi}, \citet{jinAttentionGuidedDeep2022} achieved an MAE of 3.03. Facial images have also been employed for BMI prediction, achieving an MAE of 2.3 on the VIP-attribute dataset~\cite{vip_attribute}.
In addition, regarding full-body images, work by~\citet{kimMultiViewBodyImageBased2023} explored the effect of combining three different perspectives (front, side, and back) in a single model for BMI estimation, achieving an MAE of 3.12 on their own dataset.

%% file: 03_method.tex
\section{Methodology}
\label{sec:method}

In this section, we present our approach for estimating BMI from real-world smartphone camera images. We first describe the curation of the WayBED dataset, introducing our procedure for excluding unsuitable images from the real-world dataset (see Section~\ref{sec:methods:image_filtering}). Next, we describe our computer vision model for BMI estimation in Section~\ref{sec:methods:deep_learning_model}. Finally, we describe our experiments and evaluation strategy in Section~\ref{sec:methods:evaluation}.

\subsection{Real-world dataset curation: WayBetter BMI Estimation Dataset~(WayBED)}
\label{sec:methods:image_filtering}
The dataset was collected as part of the WayBetter weight loss program, in which participants were instructed to capture front-facing, full-body photographs at multiple stages of their fitness journey, primarily for progress tracking purposes. A subset of all participants consented for their data to be used in further research, as we describe in Section~\ref{sec:experiments:dataset}.

\paragraph{Quality assurance}
Along with each photo, users were asked to report their current weight, while height was self-reported at the beginning of the program. For each body photo submitted, participants were also required to upload a photo of their scale displaying their weight, accompanied by a randomly assigned word written on paper, used as a timestamping mechanism to verify the timing and authenticity of submissions~("weigh-in words"). Each submission was reviewed by a team of human referees, who verified the authenticity and adherence to photographic guidelines, accepting or rejecting images as necessary. To maintain data integrity and deter cheating in the program, WayBetter implemented a system for evaluating both the submitted images and associated weigh-in-words along with a defined set of acceptance criteria for images. Further details of the dataset are provided in Section~\ref{sec:experiments:dataset}.

\paragraph{Automatic filtering steps}
Images captured in real-life scenarios often exhibit substantial variation in factors such as image quality, lighting conditions, and subject posture. Some images - particularly those taken from uncommon angles~(e.g., selfies) or featuring atypical postures - may be unsuitable for accurate BMI estimation. To ensure the quality and consistency of our dataset, we implement an automated filtering procedure to curate the collected images. This process aims to exclude images that do not (clearly) depict a human body, those captured from extreme camera angles (e.g., excessively high or low viewpoints), and images with outlier postures.

For the filtering process, we apply a series of steps based on human bounding box and body keypoint estimations, both obtained using the Detectron2 framework~\cite{detectron2}. Detectron2 provides models for human bounding box estimation, including confidence scores, as well as skeleton estimation comprising 17 body keypoints.
Using the estimated bounding boxes and keypoints, we implement three filtering steps:

\begin{enumerate}
  \item Person detection: Images with a human bounding box confidence score below a threshold are removed.
  \item Person-to-background ratio: Images with a bounding-box-to-image-area ratio below a predefined threshold are excluded, filtering out cases where the person occupies too little of the image~(see Section~\ref{sec:experiments:dataset}).
  \item Posture clustering: Images with outlier body keypoints, identified through clustering, are excluded.
\end{enumerate}
Each filtering step is applied independently, and the images that meet one or more exclusion criteria are removed.

\subsubsection{Posture clustering}
\label{sec:methods:image_filtering:pose_clostering}
For each image, we extract COCO-17 skeleton keypoints~\cite{coco_keypoints} and normalize them by the image’s width and height to account for variations in scale and position. This results in 34 normalized values per image (17 keypoints with x and y coordinates).

Since not all keypoints contribute equally to assessing pose quality, we apply Principal Component Analysis (PCA) to reduce redundancy and emphasize the most informative patterns. We retain the principal components that preserve 95\% of the total variance, thereby reducing noise while maintaining relevant pose information. Subsequently, we perform K-means clustering on the PCA-transformed data, testing values of k from 1 to 10, and selecting the optimal k using the elbow method. This clustering groups similar poses and enables the identification of undesirable ones—such as distorted limb positions or elevated camera angles (e.g., selfies). Finally, members of WayBetter visually inspected each image cluster to identify the types of images it represented, retaining clusters with suitable poses and discarding those deemed unsuitable, e.g., selfies.

\subsection{Mobile-capable computer vision model for BMI estimation}
\label{sec:methods:deep_learning_model}
\begin{figure}[h!t]
\centering
\includegraphics[width=\linewidth]{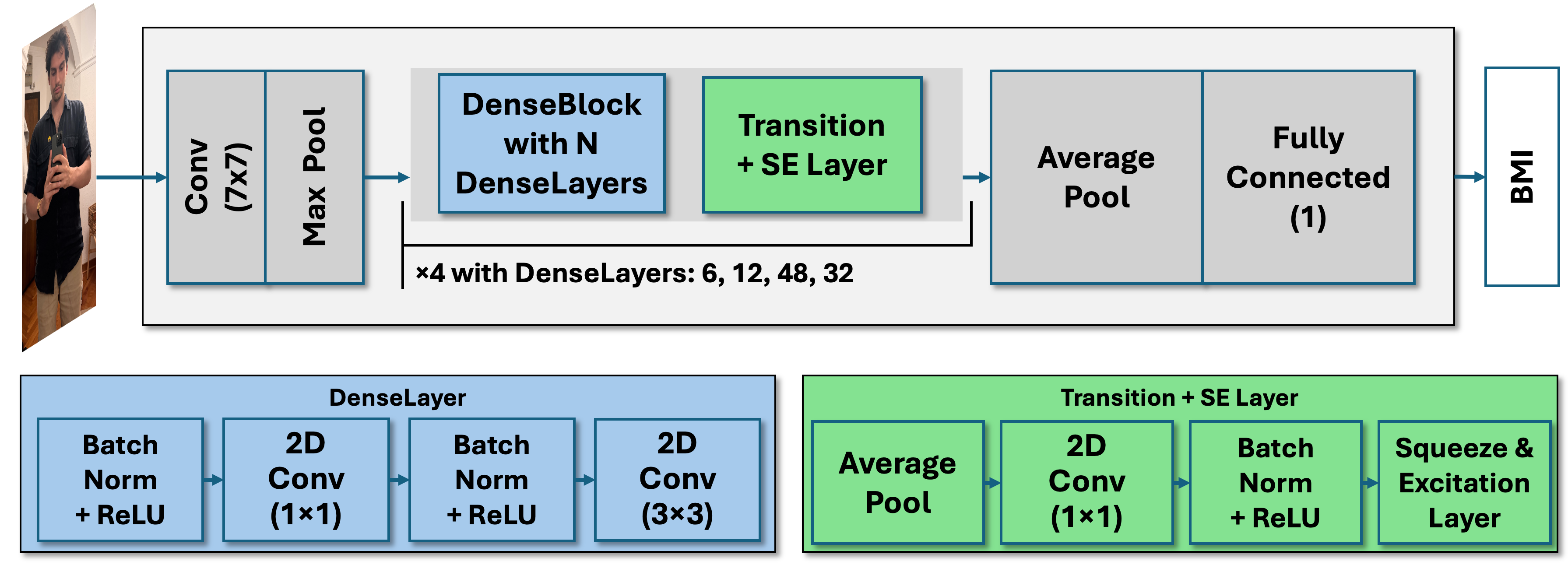}
\caption[caption]{DenseNet-201 architecture with Squeeze-and-Excitation blocks adapted from~\citet{jinAttentionGuidedDeep2022}.}
\label{fig:model_architecture}
\end{figure}

We adopt the architecture proposed by \citet{jinAttentionGuidedDeep2022} for BMI estimation from 2D body images. The model is built upon a DenseNet CNN \cite{densenet}. \citet{jinAttentionGuidedDeep2022} demonstrated that this architecture yields superior performance compared to other architectures in BMI prediction tasks. While their implementation used a DenseNet-121 backbone, we utilize DenseNet-201 to leverage its increased capacity for learning complex features from larger datasets. An overview of the architecture is shown in Figure~\ref{fig:model_architecture}.

\subsubsection{Attention mechanism}

Following \citet{jinAttentionGuidedDeep2022}, the architecture incorporates Squeeze-and-Excitation (SE) blocks~\cite{squeeze_and_excitation} after each transition layer. These blocks implement channel-wise attention by first compressing spatial information through global average pooling and then applying two 1×1 convolutions, separated by a bottleneck layer, to compute channel weights. The resulting weights are used to rescale the feature maps, allowing the model to prioritize more informative channels. For additional details, we refer to \citet{jinAttentionGuidedDeep2022}.

\subsubsection{Mobile deployment}
To enable BMI estimation in real-world settings, we deploy the full pipeline, from image capture to filtering and inference, directly on mobile devices. To compose and manage this pipeline on-device, we use the CLAID framework~\cite{claid}, which facilitates the loading, configuration, and combination of modular processing components. We employ existing CLAID modules for image capture and preprocessing, and introduce three custom modules for subsequent image analysis:
\begin{enumerate}
\item PersonFilter: Filters bounding boxes to ensure a sufficient person detection confidence and an adequate person-to-background ratio.
\item PoseFilter: Removes outlier poses based on body keypoints.
\item BMIEstimator: Applies the trained BMI estimation model locally on filtered images.
\end{enumerate}

The BMIEstimator executes our BMI estimation model on-device. We reimplemented the model proposed by \citet{jinAttentionGuidedDeep2022} in PyTorch~\cite{pytorch} and converted it for mobile inference using ExecuTorch. PersonFilter and PoseFilter use EfficientDet Lite4 and MoveNet Thunder models, respectively, based on TensorFlow Lite.

\begin{figure}[h!t]
\centering
\includegraphics[width=1.0\linewidth]{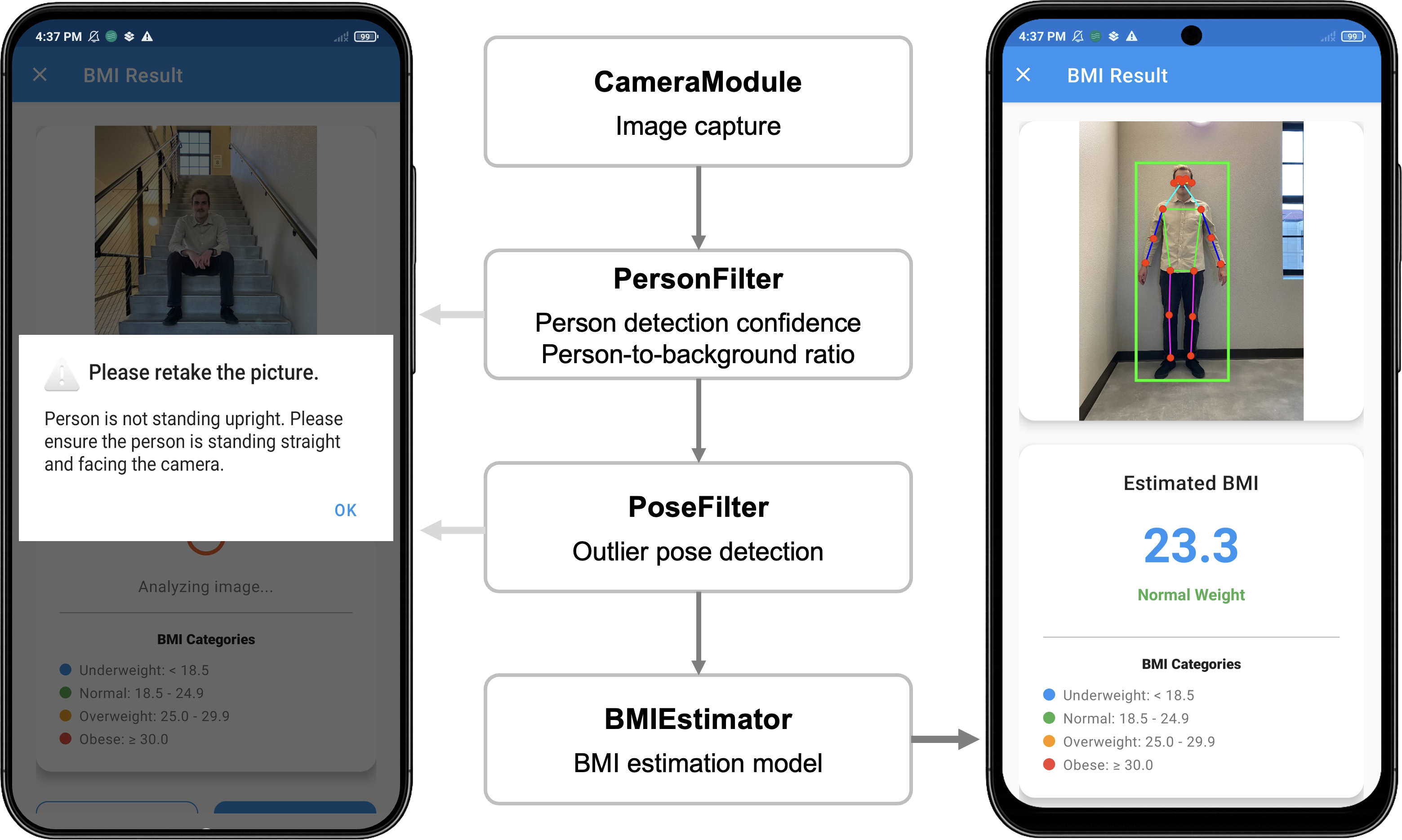}
\caption{BMI estimation mobile App.}
\label{fig:mobile_app}
\end{figure}
To facilitate reuse in custom applications, we packaged our model and inference code as a CLAID module~\cite{claid}. Figure~\ref{fig:mobile_app} illustrates our BMI estimation app, showcasing an example result.

\subsection{Evaluation}
\label{sec:methods:evaluation}
To derive a robust machine learning model for BMI estimation suitable for real-world deployment, we follow a multi-step evaluation procedure. We first assess the effectiveness of different image perspectives~(full-body, torso-up, face-only) on the WayBED dataset to identify which yields the most accurate predictions. Next, we then evaluate the generalization ability of the best-performing model and perspective on the unseen Visual-Body-To-BMI dataset, which includes new individuals. Since real-world deployment often entails distribution shifts, we further investigate whether our pretrained model—trained on the large WayBED dataset—can be effectively finetuned on a small subset of Visual-Body-To-BMI for domain adaptation. 

\subsubsection{Performance with different image perspectives on WayBED dataset}
\begin{figure}[h]
  \centering
  \includegraphics[width=\linewidth]{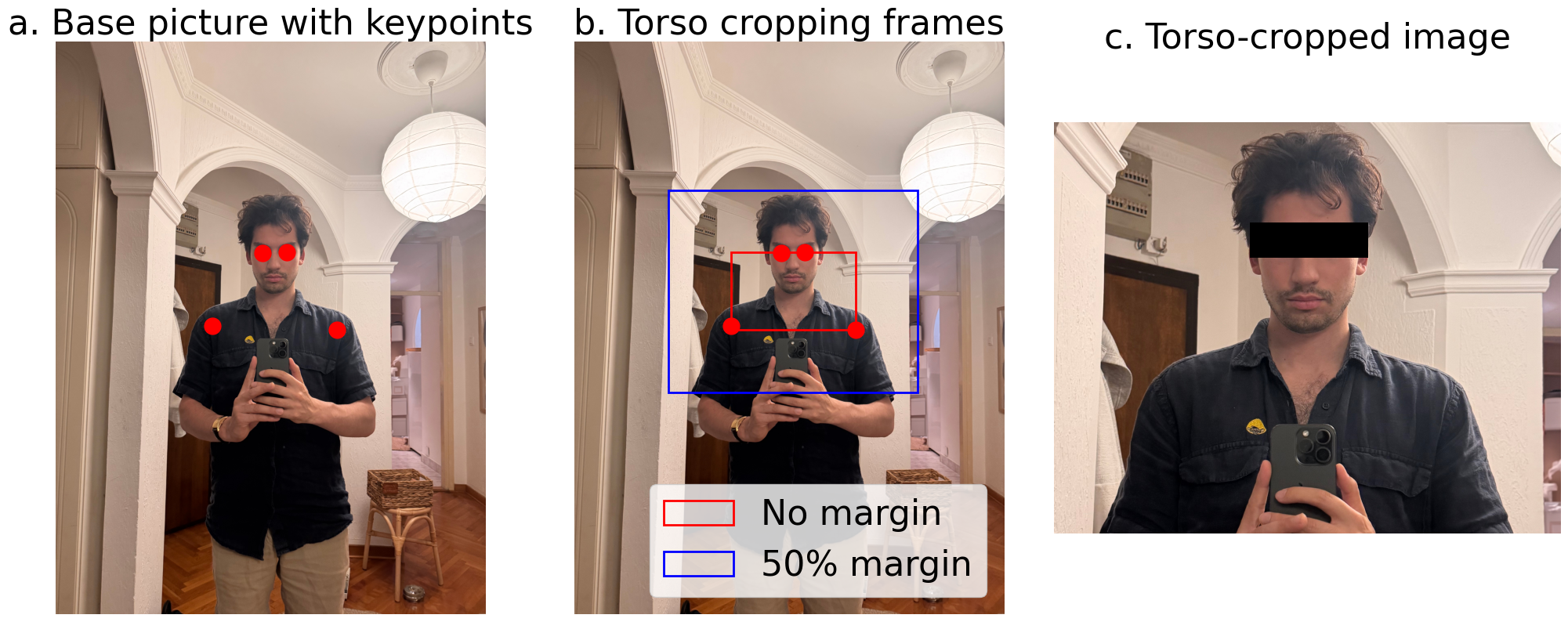}
  \caption[caption]{Example of torso-up cropping strategy~(private photo with consent).}
  \label{fig:torso_crop_demo}
\end{figure}

To identify the most suitable perspective for BMI estimation among those commonly used in the literature, i.e., full-body, torso-up, and face-only, we extract all three crops from the original full-body image for each individual in the dataset. We divide the dataset into training, validation, and test sets in a 70/15/15 ratio. To ensure subject-disjoint splits, we group images by individuals. The users are randomly shuffled and then sequentially assigned to the split (training, validation, or test) that currently has the largest gap between its target and actual number of images. Importantly, the individuals per splits are kept identical across all perspectives to ensure a fair comparison using the same set of test subjects. 
We generate face-only crops using the RetinaFace library~\cite{retinaface2020}. For torso-up crops, we use keypoint annotations obtained during the image filtering process (see Section~\ref{sec:methods:image_filtering:pose_clostering}), specifically shoulder and eye coordinates. The crop width is defined as the horizontal distance between the shoulders, and the height spans vertically from the lowest shoulder to the highest eye keypoint.  To ensure full coverage of torso and head, we add a margin equal to 50\% of the crop’s width and height. Figure~\ref{fig:torso_crop_demo} shows examples for each perspective derived from one image.

\subsubsection{Evaluation on unseen Visual-Body-To-BMI dataset}
Using the best-performing image perspective, we evaluate our best-performing model \textbf{trained on the WayBED dataset} on the \textbf{unseen Visual-Body-To-BMI dataset}~\cite{visual_body_to_bmi} to assess generalization to unseen data. The Visual-Body-To-BMI dataset consists of 5900 images, and we use 30\% for the evaluation following prior work~\cite{jinAttentionGuidedDeep2022}.

\subsubsection{Finetuning on VisualBodyToBMI}
Real-world deployment may involve distribution shifts, e.g., new individuals, environments, or image conditions relative to the training data. To maintain performance without full retraining, the model must be adaptable to new datasets. We evaluate this by finetuning our pretrained model on new data using three strategies: (1) fully unfrozen, all layers updated; (2) partially unfrozen, starting from the last dense block; and (3) only the last layer is retrained.

%% file: 04_experiments.tex
\section{Results}
\label{sec:experiments}

In this section, we present the results of our experiments. Section~\ref{sec:experiments:dataset} provides an overview of the WayBED dataset and its curation. In Section~\ref{sec:experiments:results}, we report model performance across different image perspectives (full-body, torso-up, and face-only). Finally, Section~\ref{sec:results:full_body_vbmi} presents the model’s generalization performance on the VisualBodyToBMI dataset using the best-performing perspective.

\subsection{WayBED dataset preparation}
\label{sec:experiments:dataset}
The WayBED dataset contains 84,963 images from 25,353 individuals, collected over more than a decade via the WayBetter fitness app. Images were captured by users on a wide range of Android and iOS devices. In total, over 900,000 users submitted more than 5 million images, of which 25,353 individuals  provided explicit consent for research use under the condition that their images would not be publicly released and could not be reverse-engineered or inferred from any published results. The dataset includes 49\% self-identified female and 9\% male participants; 42\% did not report gender. Figure~\ref{fig:descriptive_stats} (appendix) shows the distributions of weight, height, and BMI.

\subsubsection{Filtering} 
\label{sec:experiments:filtering}
\begin{figure}[h!]
  \centering
  \includegraphics[width=1.0\linewidth]{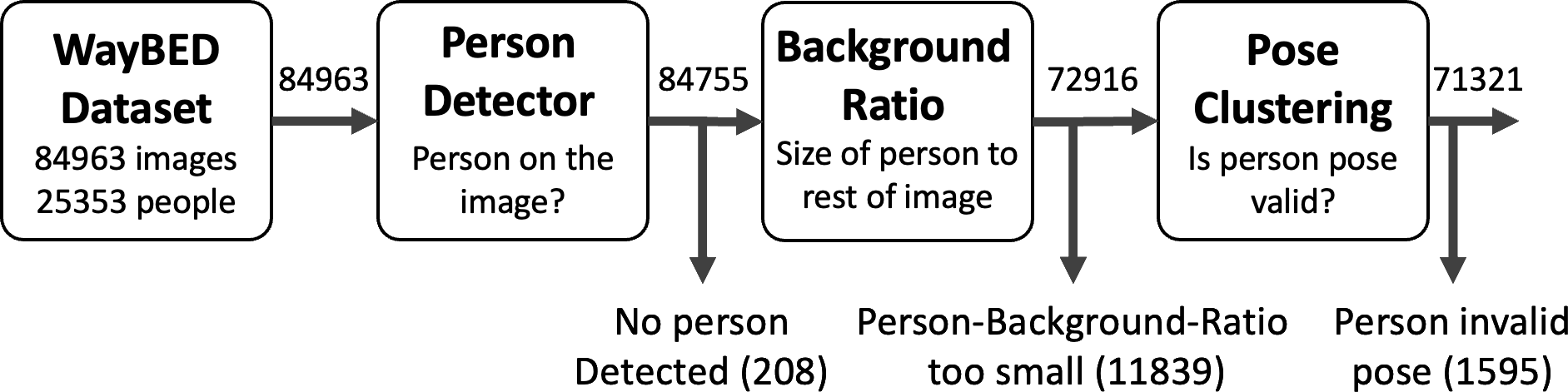}
  \caption{Images filtered out by each step.}
  \label{fig:filter_venn_diagram}
\end{figure}

\begin{figure}[h]
  \centering
  \includegraphics[width=0.8\linewidth]{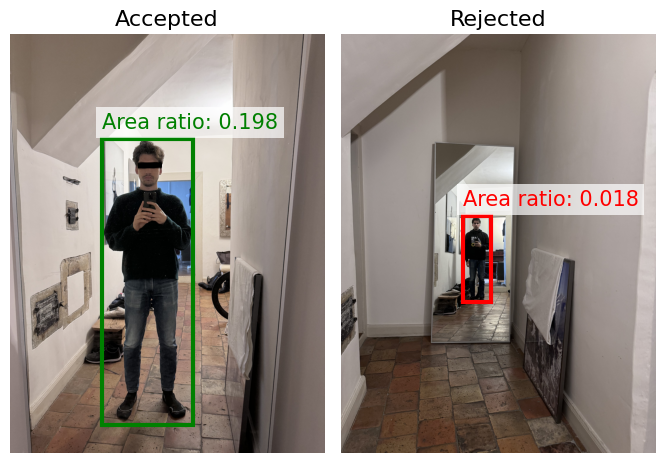}
  \caption[caption]{Example of person-to-background ratio filtering~(private photo with consent)}
  \label{fig:area_ratio_filter_demo}
\end{figure}

Applying the filtering procedure from Section~\ref{sec:methods:image_filtering}, 13,642 images (16.2\%) were removed prior to training. Specifically, 12,402 images were excluded due to too small bounding boxes, 513 due to low bounding box confidence, and 2,052 due to unsuitable poses.  Figure~\ref{fig:filter_venn_diagram} visualizes the filtering process, and Figure~\ref{fig:area_ratio_filter_demo} shows an image being rejected due to the person being too small.

\subsubsection*{Posture clustering}

The optimal number of clusters was determined to be 4 using the elbow criterion (see Figure~\ref{fig:k_means_elbow_plot} in the appendix). Clusters 1 and 2, comprising 43,192 and 39,719 images respectively, depict individuals standing upright in a frontal pose—distinguished primarily by arm position (arms by the sides vs. raised arms to hold a camera). Clusters 3 and 4, containing 2,051 and 1 image, represent outlying poses and a non-human image, often with alternative camera angles. These two clusters were excluded from the dataset. Figure~\ref{fig:cluster_variance_ellipses} in the Appendix shows the mean position and variance of the key points for clusters 1, 2, and 3. Between clusters 1 and 2, the difference in arm position is visible in the positioning of the wrist key points. Cluster 3 shows no discernible human pose. Figure~\ref{fig:clustering_umap_results} (appendix) illustrates cluster separation using a UMAP projection~\cite{umap}.

\subsection{Estimation performance with different perspectives on WayBED dataset}
\label{sec:experiments:results}

Table~\ref{tab:results_waybetter} reports MAPE and MAE for different image perspectives on the WayBetter dataset. Full-body perspectives achieve the lowest BMI error (MAPE 7.9\%, MAE 2.56), followed by torso-up (9.1\%, 2.97) and face-only (11.1\%, 3.66). The same pattern holds for absolute weight error, with full-body perspective yielding the lowest MAE (7.29~kg), compared to torso-up (8.41~kg) and face-only (10.38~kg).

\begin{table}[h!t]
  \caption{
    WayBED BMI estimation results by perspective.
  }
  \centering
    \footnotesize
    \begin{tabular}{lccc}
      & Full-Body & Torso-Up & Face-Only \\
      \midrule
      MAPE (\%) & \textbf{7.9} & 9.1 & 11.1 \\
      MAE (BMI) & \textbf{2.56} & 2.97 & 3.66 \\
      MAE (kg) & \textbf{7.29} & 8.41  & 10.38     \end{tabular}
  \vspace{-0.05in}
  \label{tab:results_waybetter}
\end{table}

\subsection{Performance analysis of full-body images}
\label{sec:results:full_body_vbmi}

Table~\ref{tab:results:models_and_datasets} reports MAPE for BMI estimation using full-body (best perspective) images with DenseNet-121 and DenseNet-201 across the training and evaluation settings described in Section~\ref{sec:methods:evaluation}. When trained and evaluated on WayBED, DenseNet-201 achieved a MAPE of 7.90\%. Training only on WayBED and evaluating on the unseen test-set of VisualBodyToBMI yielded higher errors for both models, with DenseNet-201 reaching 13.38\%.  When trained and evaluated on VisualBodyToBMI, DenseNet-121 reached 9.28\% MAPE and DenseNet-201. DenseNet-201 achieved 8.75\%. Fine-tuning DenseNet-201 on VisualBodyToBMI reduced the error to 8.56\%. 

\begin{table}[h!]
\centering
\caption{Results of BMI estimation.}
\footnotesize

\label{tab:results:models_and_datasets}
\begin{tabularx}{\linewidth}{p{1.85cm}p{2.55cm}XX}
\textbf{Model} & \textbf{Trained on} & \textbf{Eval on} & \textbf{MAPE} \\
\midrule
DenseNet-121 & WayBED & WayBED & 11.7 \\
DenseNet-121 & WayBED & VBMI & 23.10 \\
DenseNet-121 & VBMI & VBMI & 9.28 \\
DenseNet-121 & WayBED, ft. VBMI & VBMI & 10.72 \\ 
\midrule
DenseNet-201 & WayBED & WayBED & \textbf{7.90} \\
DenseNet-201 & WayBED & VBMI & 13.38 \\
DenseNet-201 & VBMI & VBMI & 8.75 \\
DenseNet-201 & WayBED, ft. VBMI & VBMI & \textbf{8.56} \\
\end{tabularx}
\caption*{\footnotesize Note: "ft. VBMI" means finetuned on VisualBodyToBMI; finetuning results are for fully unfrozen model; see Appendix~\ref{sec:appendix:finetuning}.}

\end{table}
\vspace{-5mm}

%% file: 05_discussion.tex
\section{Discussion}
\label{sec:discussion}
In this section, we discuss the practical implications of our findings and outline the study’s limitations.


\paragraph{Principal findings}

Our model shows strong generalization to unseen data, achieving a MAPE of 13.38\% on the test portion of the VisualBodyToBMI dataset~(full-body), despite not being included during training. 
On WayBED, full-body images yielded the best performance (MAPE 7.9\%), highlighting the importance of context and background for BMI estimation.
The torso-up perspective also performed well (MAPE 9.1\%), suggesting it may serve as a viable alternative when full-body images are unavailable. Face-only inputs resulted in the lowest performance (MAPE 11.1\%).

\paragraph{Comparison to prior work}
To the best of our knowledge, we present the first evaluation of BMI estimation on a large-scale, real-world dataset comprising 84,963 images—substantially larger than prior datasets, which included 14,500 facial or 5,900 full-body images (e.g., VisualBodyToBMI~\cite{visual_body_to_bmi}). Without training on Visual-Body-To-BMI, our model achieves a MAPE of 13.38\%, closely matching the original 12.50\%~\cite{visual_body_to_bmi} reported with in-dataset training. Fine-tuning on VisualBodyToBMI after training on WayBED reduces the MAPE to 8.56\%, setting a new state of the art. This also constitutes the first cross-dataset evaluation of a pretrained BMI model on an entirely unseen dataset. Additionally, we present the first direct comparison of facial and full-body perspectives within a single dataset, with our full-body model achieving a MAPE of 7.9\%, improving over the previously reported 9.6\% by \citet{jinAttentionGuidedDeep2022}.  

\paragraph{Practical implications}
A MAPE of 7.9\% in BMI estimation aligns with similar tools for assessing body composition and body fat, where a MAPE between 6–8\% is generally considered good~\cite{nature_body_fat_percentage_estimation}. This performance suggests potential for applications where direct measurements are impractical, such as large-scale population studies, where individuals tend to underreport their BMI~\cite{bmi_underreporting}, or telehealth applications. While the accuracy declines for torso-only or face-only perspectives, these may still hold value in telehealth contexts to enable BMI estimation during video consultations. For clinical use cases, however, the achieved MAE of 2.56 BMI might lead to individuals being misclassified across critical BMI thresholds (e.g., shifting from “overweight” to “obese”), which might trigger clinical evaluation or interventions~\cite{bmi_intervention_cutoff_points}.  BMI estimation for clinical applications should be as accurate as possible~\cite{when_is_estimation_good_enough} and thus requires further research. To support such efforts, we fully release the code for replicating and deploying our approach. While the WayBED-trained model cannot be publicly released due to privacy requirements, we may share it upon request. We also provide code for training and deploying models on custom datasets. All models and filtering modules for rejecting unsuitable images are available via the CLAID~\cite{claid} framework for mobile deployment.

\paragraph{Limitations}
Based on our experiments, we identify the following limitations of our study:

\begin{enumerate}
  \item \textit{Estimation from a single image:} Our model estimates BMI from a single image; incorporating images from different perspectives or time points could reduce error, particularly for tracking relative BMI changes.
  \item \textit{Lack of scale reference:} Images were captured at varying distances, resulting in inconsistent scales of the person within each image. Introducing reference objects (e.g., a paper sheet or credit card) during image capture could potentially reduce estimation error.
  \item \textit{Self-reported height and weight:} Labels were self-reported by participants, which may introduce measurement errors and reporting biases.
  \item \textit{Gender imbalance:} The dataset shows a gender imbalance: 49\% female, 9\% male, and 42\% not reported, which may affect performance across genders.

\end{enumerate}

%% file: 12_appendix.tex
\section*{Appendix}
\label{sec:appendix_section}

\subsubsection{Dataset statistics}
\begin{figure}[h!t]
  \centering
  \includegraphics[width=\linewidth]{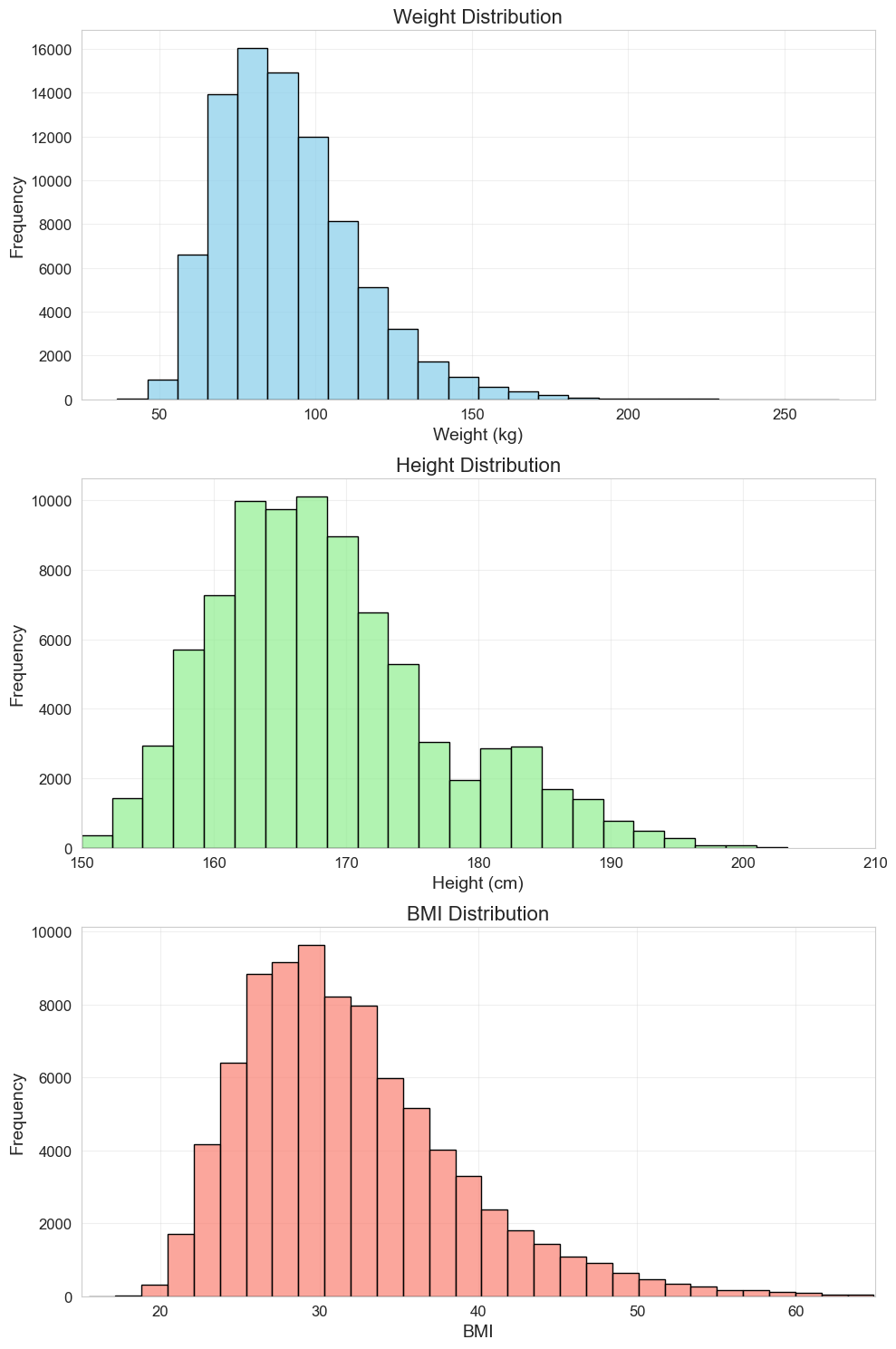}
  \caption{WayBED dataset statistics.}
  \label{fig:descriptive_stats}
\end{figure}

Figure~\ref{fig:descriptive_stats} provides statistics for the WayBED dataset, showing the distribution of height, weight and BMI over all individuals. 


\subsubsection{Pose clusters}
\label{sec:appendix:dataset_filtering}
\begin{figure}[h!t]
  \centering
  \begin{subfigure}[b]{0.65\linewidth}
    \includegraphics[width=\linewidth]{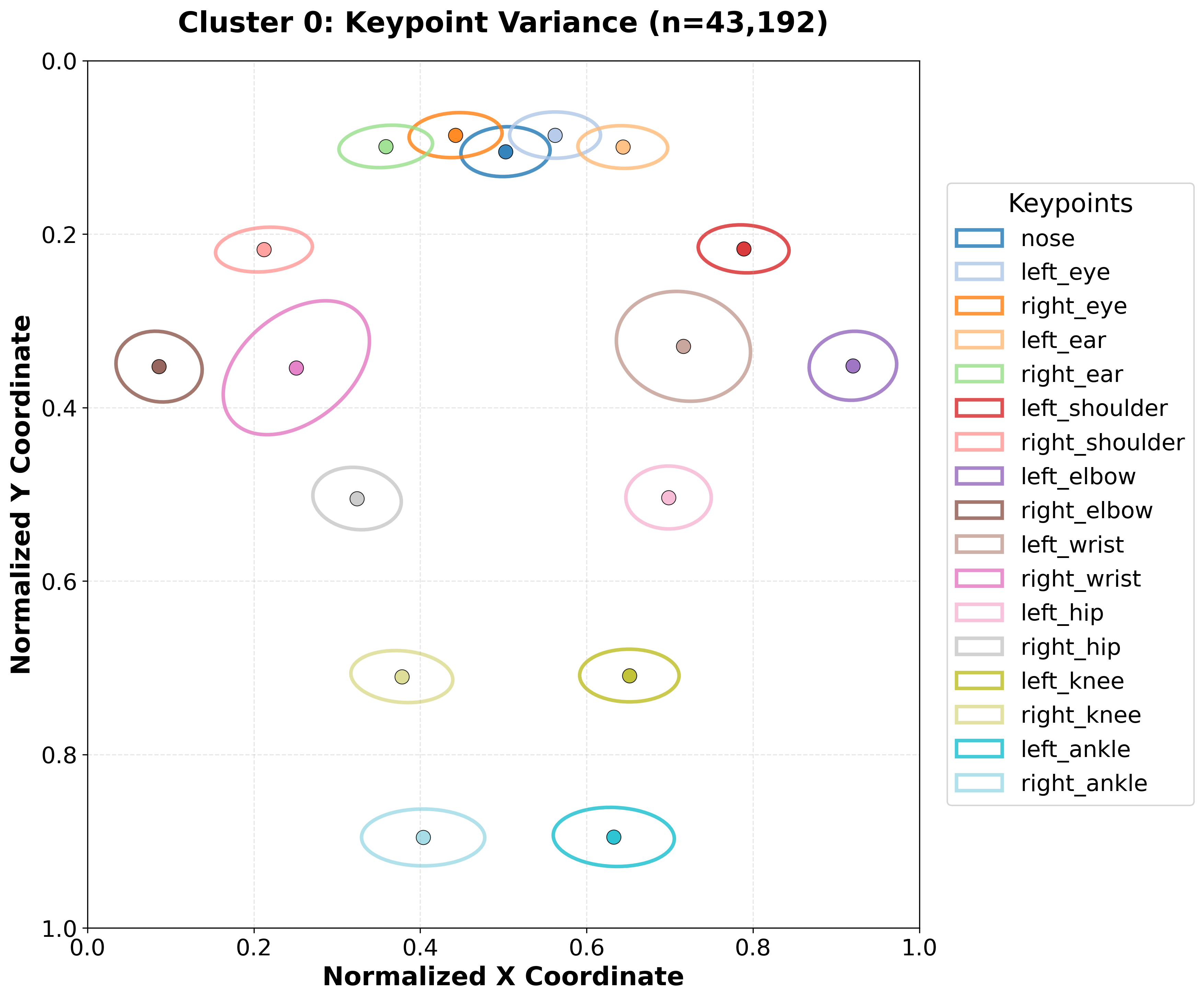}
    \caption{Cluster 0}
  \end{subfigure}\hfill%
  \begin{subfigure}[b]{0.65\linewidth}
    \includegraphics[width=\linewidth]{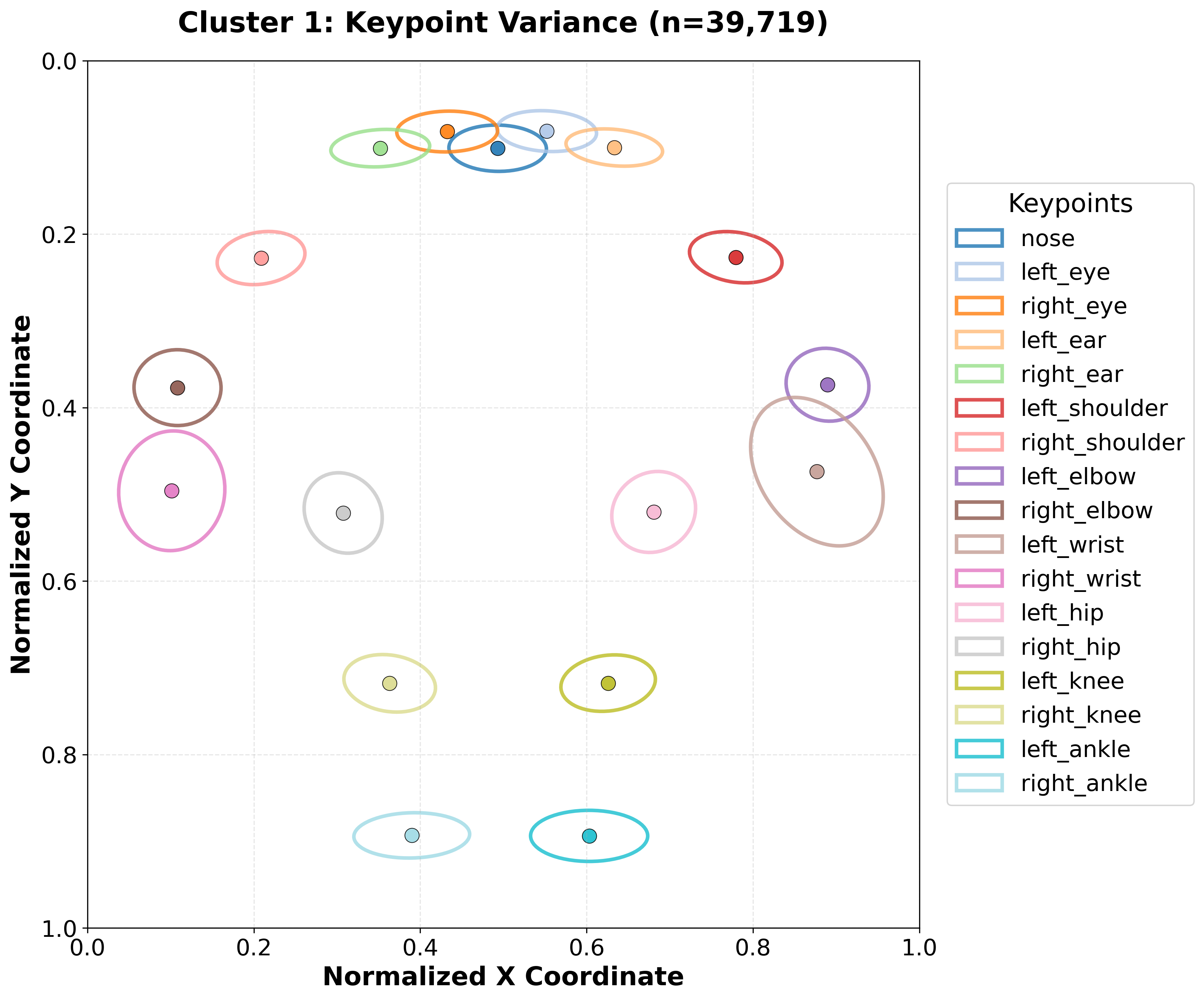}
    \caption{Cluster 1}
  \end{subfigure}\hfill%
  \begin{subfigure}[b]{0.65\linewidth}
    \includegraphics[width=\linewidth]{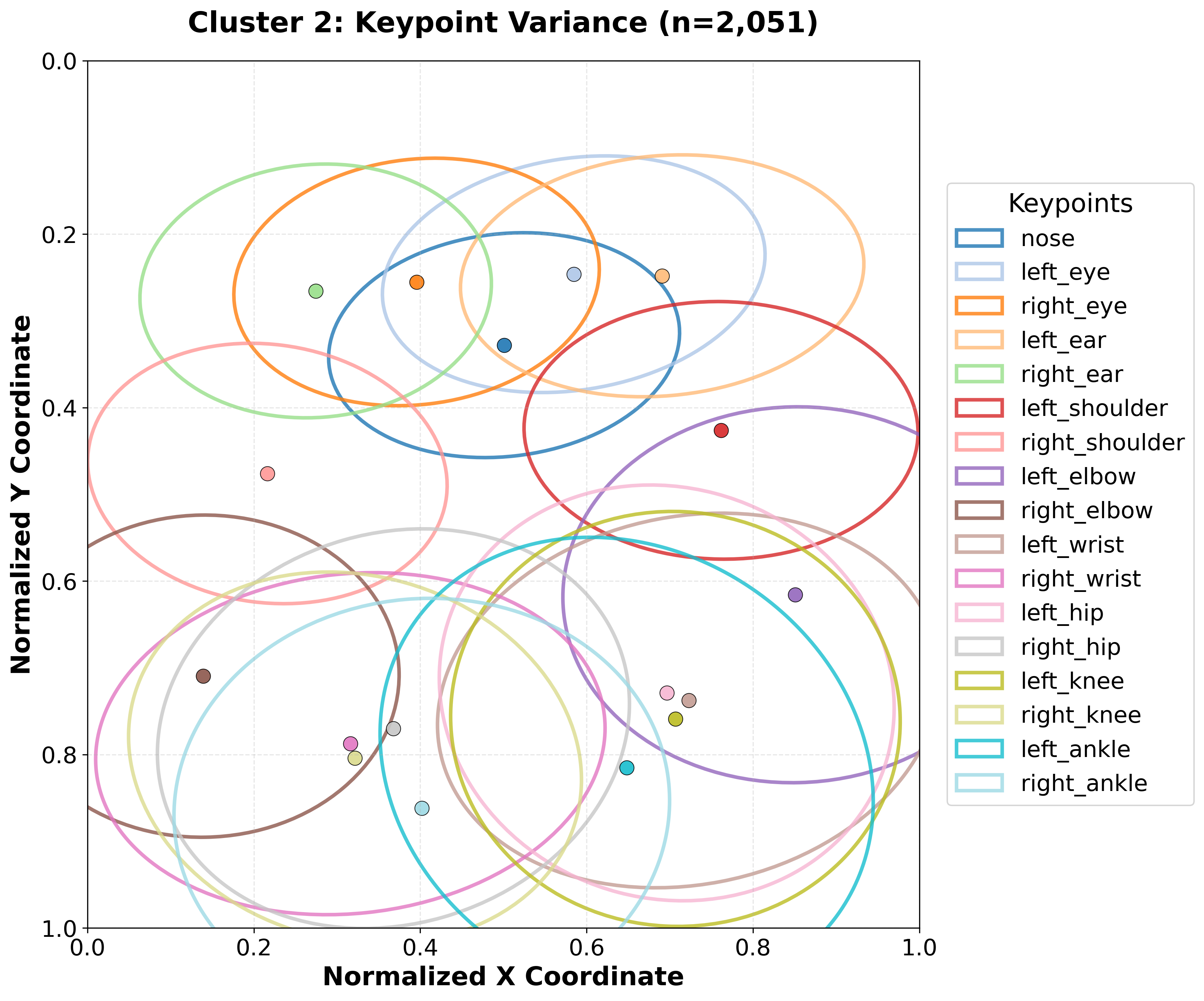}
    \caption{Cluster 2}
  \end{subfigure}
  \caption{Posture clusters with keypoint variance ellipses.}
  \label{fig:cluster_variance_ellipses}
\end{figure}

To filter out images containing invalid poses, such as selfies or extreme close-ups, we implemented a posture-based clustering mechanism. Details of this method are provided in Section~\ref{sec:methods:image_filtering:pose_clostering}. 
To derive these clusters, we first performed skeleton pose estimation on all images in the dataset.
Figure~\ref{fig:cluster_variance_ellipses} illustrates the variance of detected body keypoints across all pictures, highlighting the common positions of corresponding joints. 

\begin{figure}[h!t]
\centering
\includegraphics[width=\linewidth]{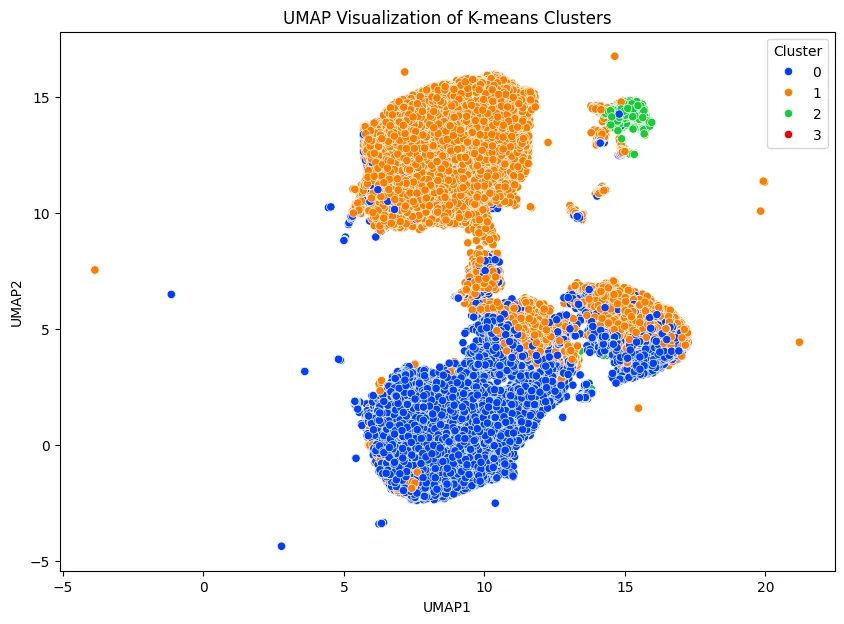}
\caption{UMAP plot for $k=4$ clusters of body keypoint data.}
\label{fig:clustering_umap_results}
\end{figure}

\begin{figure}[h!t]
\centering
\includegraphics[width=\linewidth]{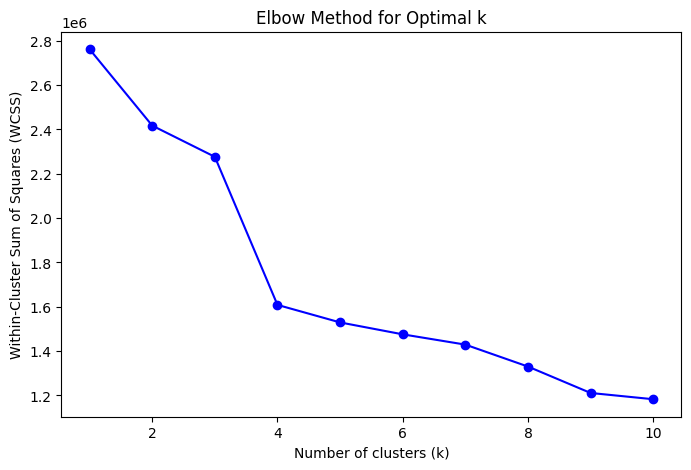}
\caption{Elbow plot for k-means clustering of body keypoint data.}
\label{fig:k_means_elbow_plot}
\end{figure}

After detecting the keypoints, we perform a PCA as detailed in Section~\ref{sec:methods:image_filtering:pose_clostering} and use $k$-means to cluster the data. Figure~\ref{fig:clustering_umap_results} presents a UMAP visualization of the resulting clusters after dimensionality reduction via PCA, while Figure~\ref{fig:k_means_elbow_plot} shows the within-cluster variance for different values of $k$ in the $k$-means algorithm.

\subsubsection{Training details}
We train each model for 40 epochs with a batch size of 64, using the Adam optimizer \cite{adam_optimzer} with a learning rate of 0.001 and a weight decay of 0.0001. The models are optimized using the Mean Squared Error (MSE) loss function, following the setup used by Jin et al.~(2022) \cite{jinAttentionGuidedDeep2022}. We monitor MAE and MAPE at each epoch for evaluation purposes. No additional hyperparameter tuning is performed.
To promote convergence and reduce the risk of overfitting, we apply dynamic learning rate scheduling: the learning rate is reduced by a factor of 0.1 if the validation loss does not improve for 5 consecutive epochs. All models are implemented in PyTorch \cite{pytorch} and trained on a single NVIDIA T4 GPU.

\begin{figure*}[h!t]
\centering
\includegraphics[width=1\textwidth]{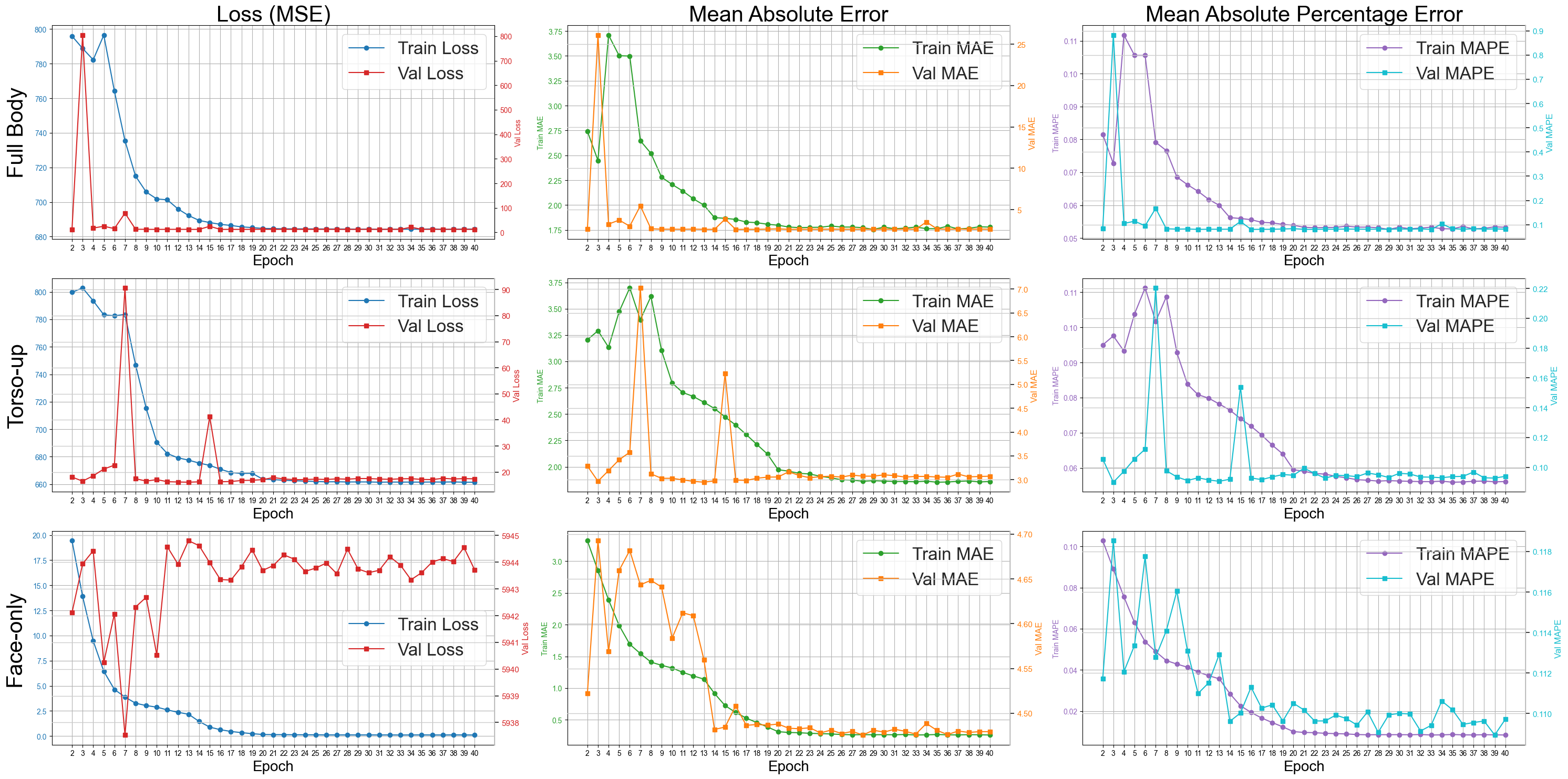}
\caption{Loss, MAE and MAPE per epoch for the training and validation sets of each model.}
\label{fig:training_metrics}
\end{figure*}

\subsubsection{Finetuning results}
\label{sec:appendix:finetuning}

Table~\ref{tab:densenet_finetune} shows classification error rates for DenseNet-121 and DenseNet-201 for different fine-tuning strategies, when training the models on WayBED and finetuning on VisualBodyToBMI. Unfreezing only the classifier yields errors of 16.07\%~(DenseNet-121) and 13.04\%~(DenseNet-201). Unfreezing the last dense block reduces errors to 12.40\% and 9.38\%, respectively. Fully unfreezing all layers achieves the lowest errors: 10.72\% for DenseNet-121 and 8.56\% for DenseNet-201. Error consistently decreases as more layers are unfrozen.

\begin{table}[h!]
\centering
\begin{tabularx}{\linewidth}{lXX}
\toprule
\textbf{Fine-Tuning Strategy} & \textbf{DenseNet-121} & \textbf{DenseNet-201} \\
\midrule
Unfreeze classifier       & 16.07\% & 13.04\% \\
Unfreeze last dense block & 12.40\% & 9.38\%  \\
Unfreeze all              & 10.72\% & 8.56\%  \\
\bottomrule
\end{tabularx}
\caption{Classification error rates for DenseNet-121 and DenseNet-201 under different fine-tuning strategies.}
\label{tab:densenet_finetune}
\end{table}